\PassOptionsToPackage{dvipsnames}{xcolor}

\documentclass{Interspeech}
\usepackage{amsmath}
\usepackage{colortbl}
\usepackage{xcolor}
\usepackage[dvipsnames]{xcolor}
\usepackage{float}  
\usepackage{flafter} 
\usepackage{hyperref}

\hypersetup{
    colorlinks=true,
    linkcolor=blue,
    filecolor=magenta,      
    urlcolor=cyan,
    pdftitle={Overleaf Example},
    pdfpagemode=FullScreen,
    }

\interspeechcameraready

\title{Voice Conversion Improves Cross-Domain Robustness \\ for Spoken Arabic Dialect Identification}

\author[affiliation={1}]{Badr M.}{Abdullah}
\author[affiliation={2}]{Matthew}{Baas}
\author[affiliation={1}]{Bernd}{Möbius}
\author[affiliation={1}]{Dietrich}{Klakow}

\affiliation{Language Science and Technology}{Saarland University}{Germany} 
\affiliation{}{Camb.AI}{UAE}
\email{\textbf{Corresponding author}  badr.nlp@gmail.com}

\usepackage{comment}

\begin{document}

\maketitle

\begin{abstract}    
Arabic dialect identification (ADI) systems are essential for large-scale data collection pipelines that enable the development of inclusive speech technologies for Arabic language varieties.
However, the reliability of current ADI systems is limited by poor generalization to out-of-domain speech.
In this paper, we present an effective approach based on voice conversion for training ADI models that achieves state-of-the-art performance and significantly improves robustness in cross-domain scenarios.
Evaluated on a newly collected real-world test set spanning four different domains, our approach yields consistent improvements of up to +34.1\% in accuracy across domains.
Furthermore, we present an analysis of our approach and demonstrate that voice conversion helps mitigate the speaker bias in the ADI dataset. 
We release our robust ADI model and cross-domain evaluation dataset to support the development of inclusive speech technologies for Arabic. 

\end{abstract}

\section{Introduction}

Arabic is the native language of more than 320 million people geographically distributed across the Middle East and North Africa \cite{brown2010concise}.
Throughout the Arabic-speaking world, Modern Standard Arabic (MSA) serves as the official language and the medium of formal communication and news broadcasts. 
However, MSA is not naturally acquired and functional knowledge of it can only be achieved through formal education. 
Arabic dialects, on the other hand, are the language varieties that Arabic speakers naturally acquire and use for daily communication. 
These dialects exhibit considerable geographic variation, with varying degrees of mutual intelligibility across regions \cite{khan2011semitic}.
Although spoken dialects are neither standardized nor formally taught, they maintain a strong cultural presence through songs, folktales, and movies \cite{habash2010introduction}.

The linguistic and regional variations of Arabic pose significant challenges for the development of Arabic speech technologies.
While modern ASR systems work well on MSA speech, they struggle with  dialectal speech \cite{talafha-etal-2024-casablanca}.
The limited dialectal resources and lack of writing standardization hinder the development of dialect-aware ASR models \cite{waheed-etal-2023-voxarabica}.
Additionally, text-to-speech systems require high-quality recordings with known sources of variation with respect to dialect and  register.
Therefore, robust Arabic dialect identification (ADI) systems are an  essential component in large-scale data collection pipelines to enable the development of inclusive speech technologies that serve a wider range of speaker communities \cite{sullivan23_interspeech}.

\begin{figure}[t]
\centering
\includegraphics[width=0.45\textwidth]{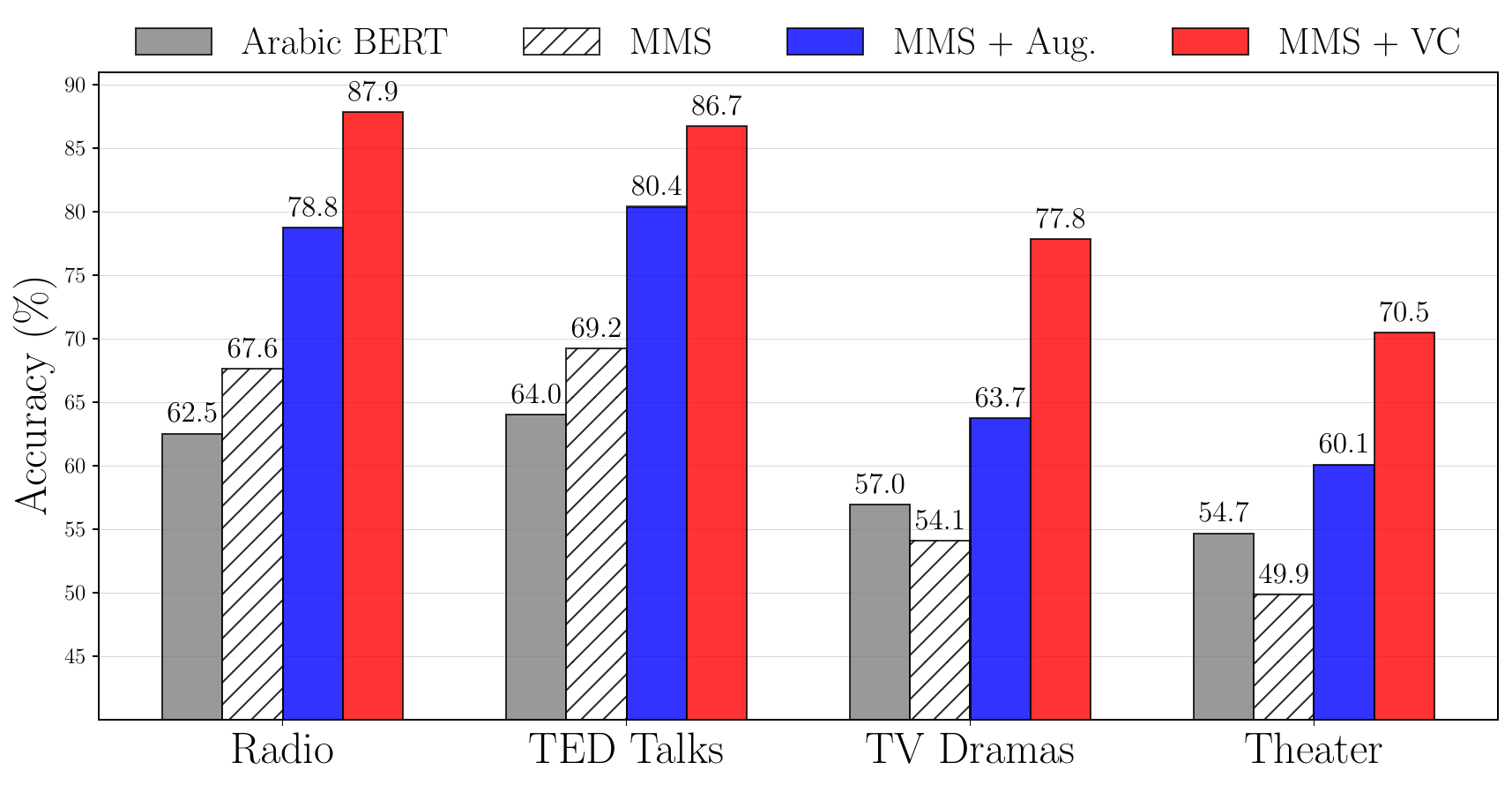}
\vspace{-8pt}
\caption{Cross-domain performance evaluation: Starting by fine-tuning the massively multilingual speech (MMS) model as a baseline, voice conversion (MMS + VC) consistently outperforms traditional data augmentation (MMS + Aug.) across four unseen domains, achieving up to +34.1\% average relative improvement in Arabic dialect identification accuracy.}
\label{fig:cross_domain_performance}
\vspace{-15pt}
\end{figure}

Although the ADI task has gained significant research interest within the speech technology community \cite{ali2017speech, shon2018convolutional, ali2019mgb, lin2020transformer}, most evaluations of ADI systems have been limited to in-domain settings where test samples are drawn from the training domain.
Recent studies have shown that while pre-trained speech models with self-supervised learning (SSL) significantly outperform conventional acoustic features and earlier neural approaches \cite{sullivan23_interspeech, kulkarni2023yet}, they still under-perform in cross-domain settings \cite{sullivan23_interspeech}.
This limitation highlights the need for training strategies that yield ADI systems that are robust to variability in recording conditions and spoken genres.
We argue that developing ADI models that transfer to unseen domains is more practical than attempting to collect diverse datasets covering all possible domains, especially since future use cases cannot be anticipated during model development.
Therefore, to reliably evaluate real-world robustness of ADI systems, models should be tested on samples from multiple unseen domains not represented in the training data---that is, a zero-shot cross-domain evaluation.
To address these challenges, we make the following contributions:

    \begin{itemize} 

        \item To evaluate cross-domain robustness of ADI systems, we create a curated multi-domain dataset of $\sim$12 hours of speech from different public sources  (§3.2).
        
        \item We propose an effective strategy  based on voice conversion for training ADI models (§\ref{section:method}) that outperforms strong baselines in both in-domain and cross-domain scenarios (§5).

        \item We further analyze our model and demonstrate that voice conversion mitigates the speaker bias in the ADI dataset (§6).
    
        \item We share our robust ADI model\footnote{\href{https://huggingface.co/badrex/mms-300m-arabic-dialect-identifier}{badrex/mms-300m-arabic-dialect-identifier}} and new evaluation dataset\footnote{\href{https://huggingface.co/datasets/badrex/MADIS5-spoken-arabic-dialects}{badrex/MADIS5-spoken-arabic-dialects}} with the research community on HuggingFace Hub.
        
    \end{itemize}

\section{Voice Conversion for ADI}
\label{section:method}

We formalize ADI as a classification problem.
Given an Arabic speech sample $\mathbf{x}$, the goal is to predict the speaker's dialect $y \in \mathcal{Y}$, where $\mathcal{Y}$ is a closed set of dialects.
To do so, we require a dataset of $N$ natural speech samples, each paired with a dialect annotation:
\begin{equation}
    \mathcal{D} = \Bigl\{ (\mathbf{x}_i, y_i)  \Bigr\}_{i = 1}^{N}
\end{equation}
The dataset $\mathcal{D}$ is used to train a model via cross-entropy loss to predict the dialect $y$ from the acoustic input $\mathbf{x}$.
In our method, we apply voice conversion (VC) to create a re-synthesized dataset from the training samples.
VC transforms a spoken utterance in a way that the generated speech is perceived as if it was spoken by a different speaker while preserving content and intelligibility \cite{mohammadi2017overview}.
Although VC has been applied to a few speech processing tasks in the literature \cite{baas2022voice, casanova2023asr, wubet2022voice}, its effectiveness for dialect identification remains unexplored.  
We formalize VC as a parametric function
\begin{equation}
    \Tilde{\mathbf{x}} = \mathcal{C}_{\mathbf{\theta}}(\mathbf{x}, \mathbf{v})
\end{equation}
where $\mathbf{v}$ is a speech sample from a target speaker, and $\Tilde{\mathbf{x}}$ is a re-synthesized segment that preserves the linguistic content of $\mathbf{x}$ but with the acoustic-phonetic features that characterizes the speaker of $\mathbf{v}$.
Modern VC techniques offer efficiency and naturalness while handling unseen speakers effectively \cite{baas23_interspeech, qian2019autovc, polyak21_interspeech, kong2020hifi}.
The transform function $\mathcal{C}_{\mathbf{\theta}}$ can be viewed as a generalized form of audio data augmentation.
For instance, SpecAugment \cite{park2019specaugment} represents a non-parametric version of $\mathcal{C}_{\mathbf{\theta}}$ that does not require a target voice.
Using VC, we obtain a re-synthesized dataset
\begin{equation}
   \Tilde{\mathcal{D}} = \Bigl\{ \bigl(\mathcal{C}_{\mathbf{\theta}}(\mathbf{x}_i, \mathbf{v}_i), y_i \bigr)  \Bigr\}_{i = 1}^{M}
\end{equation}
Here, the target voice $\mathbf{v}_i$ can be fixed for all training segments or uniformly sampled from a pool of $T$ target speakers: $\mathbf{v}_i \sim \{\mathbf{v}_1, \dots, \mathbf{v}_{T}\}$.
When each training sample is converted once, $M = N$.
Using a single target speaker yields a dataset where all training segments sound as if spoken by the same speaker.
We train our ADI model on the combined dataset $\mathcal{D}_{\text{train}}$ formed by concatenating the natural and re-synthesized datasets
\begin{equation}
    \mathcal{D}_{\text{train}} = \mathcal{D} \cup \Tilde{\mathcal{D}} = \{(\mathbf{x}_i, y_i)\}_{i=1}^N \cup \{(\Tilde{\mathbf{x}}_i, y_i)\}_{i=1}^M
\end{equation}
In our experiments, we show that VC significantly improves ADI performance both in-domain and cross-domains, achieving gains that traditional data augmentation techniques cannot match.

\section{Datasets}

\subsection{Training Dataset: MGB-3 ADI-5}
As our training dataset, we use the MGB-3 ADI-5 dataset, which is a widely-used ADI resource with coarse-grained dialect labels derived from Aljazeera TV broadcast \cite{ali2017speech}. 
It consists of approximately 14.6k samples ($\sim$53.6 hours) containing speech segments of MSA as well as four Arabic dialect groups based on geography: Gulf Arabic (spoken in the Arabian peninsula), Levantine Arabic, Maghrebi Arabic (spoken in North Africa), and Egyptian Arabic. 
The speech segments come from diverse content including news reports, panel discussions, and interviews.
The dataset is relatively balanced across dialects and features a wide range of speakers and acoustic conditions. 
The validation and test splits of ADI-5 consists of 10 hours and 10.1 hours of speech, respectively.

\definecolor{lightgray}{rgb}{0.95, 0.95, 0.95}
\definecolor{lightblue}{rgb}{0.93, 0.95, 1.0}

\begin{table*}[t]
\centering
\caption{Results comparison across different baselines and data augmentation techniques measured in accuracy (\%). The $\Delta$ columns show relative improvement over our strongest speech baseline (MMS)  trained on natural speech only. $|\mathcal{D}_{\text{train}}|$ is the size of the training dataset for each model where $N$ is the number of natural training samples in ADI-5 dataset. The results for models trained with VC are averaged across four runs with different target speaker sets. Chance-level accuracy is $\sim$20\%. }
\vspace{-8pt} 

\setlength{\tabcolsep}{4pt}
\begin{tabular*}{\textwidth}{@{\extracolsep{\fill}}llc|cc|ccccccc@{}}
\toprule
\rowcolor{white}
\multicolumn{3}{c|}{} & \multicolumn{2}{c|}{In-domain} & \multicolumn{6}{c}{Cross-domain  (MADIS-5)} \\
\cmidrule(lr){4-5} \cmidrule(lr){6-11} \rowcolor{white}
 Model & Training Data & $|\mathcal{D}_{\text{train}}|$ & \small ADI-5 & \small $\Delta$ (\%) & \small Radio & \small TEDx & \small Dramas & \small Theater & \small Avg. & \small $\Delta$ (\%) \\
\midrule
\midrule


\rowcolor{lightgray}
\multicolumn{11}{l}{\color{Blue}\textbf{Text baselines}} \\
\quad SVM & phone $n$-grams & $N$ & 66.82 & & 69.80 & 64.56 & 50.62 & 49.32 & 58.58 & \\
\quad SVM &  character $n$-grams & $N$ & 50.00 & & 55.09 & 52.56 & 47.23 & 37.20 & 48.02 & \\
\quad Arabic BERT & sub-words & $N$ & 62.73 & & 62.53 & 64.03 & 56.95 & 54.69 & 59.55 & \\
\midrule[\heavyrulewidth]

\rowcolor{lightgray}
\multicolumn{11}{l}{\color{Blue}\textbf{Speech baselines}} \\
\quad XLSR-53 & natural speech & $N$ & 68.57 & & 46.75 & 67.80 & 41.36 & 49.72 & 51.41 & \\
\quad XLSR-128 & natural speech & $N$ & 70.58 & & 52.38 & 62.67 & 48.36 & 48.76 & 53.04 & \\
\quad MMS & natural speech & $N$ & 75.94 & – & 67.63 & 69.23 & 54.12 & 49.88 & 60.22 & – \\
\midrule

\rowcolor{lightgray}
\multicolumn{11}{l}{\color{Blue}\textbf{MMS with audio augmentations}} \\
\quad MMS & natural + SpecAugment & $2 \times N$ & 78.75 & +3.70 & 70.23 & 77.60 & 56.50 & 60.81 & 66.29 & +10.08 \\
\quad MMS & natural + Pitch Shift & $2 \times N$ & 80.63 & +6.18 & 70.45 & 76.70 & 57.97 & 61.37 & 66.62 & +10.64 \\
\quad MMS & natural + RIR & $2 \times N$ & 77.88 & +2.55 & 71.82 & 76.40 & 52.32 & 50.20 & 62.69 & +04.10 \\
\quad MMS & natural + Additive Noise & $2 \times N$ & 79.42 & +4.58 & 77.31 & 78.66 & 51.30 & 59.78 & 66.76 & +10.87 \\
\quad MMS & natural + All Augmentations & $5 \times N$ & 81.64 & +7.51 & 78.76 & 80.39 & 63.73 & 60.10 & 70.75 & +17.49 \\
\midrule

\rowcolor{lightgray}
\multicolumn{11}{l}{{\color{Blue}\textbf{MMS with voice conversion (VC)}}} \\
\quad MMS & natural + VC with  1 voice  & $2 \times N$ & 82.17 & +8.21 & 77.68 & 82.79 & 67.80 & 62.62 & 72.72 & +20.77 \\
\quad MMS & natural + VC  with  2 voices & $3 \times N$ & 84.28 & +10.99 & 85.17 & 83.84 & 72.51 & 65.06 & 76.65 & +27.29 \\
\rowcolor{lightblue}
\quad MMS & natural + VC with  4 voices & $5 \times N$ & \textbf{85.32} & \textbf{+12.35} & \textbf{87.86} & \textbf{86.73} & \textbf{77.85} & \textbf{70.47} & \textbf{80.73} & \textbf{+34.07} \\
\bottomrule
\end{tabular*}

\label{tab:results}
\vspace{-12pt} 
\end{table*}

\begin{figure}[t]
\centering
\includegraphics[width=0.45\textwidth]{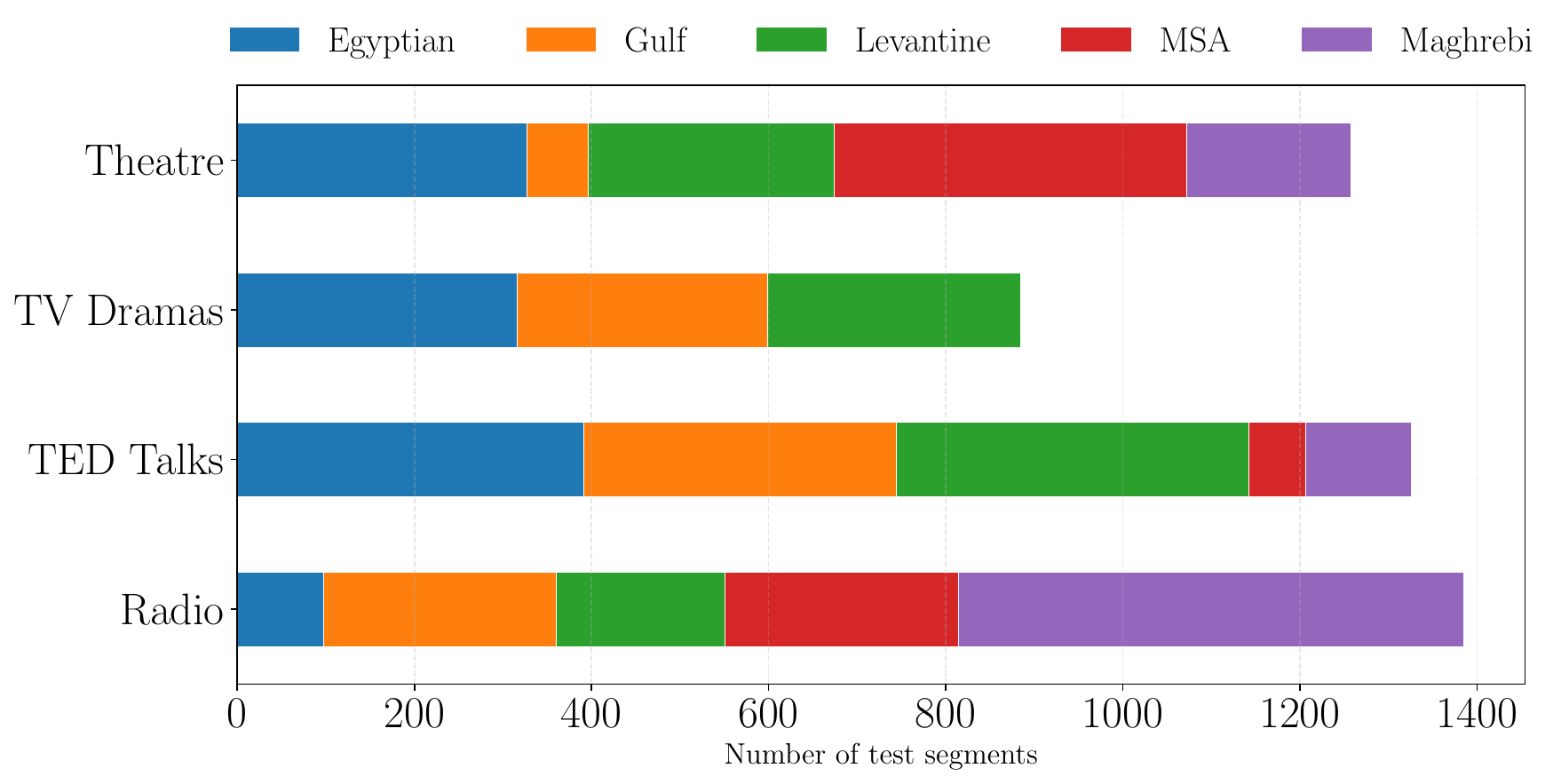}
\caption{Dialect distribution across the domains of MADIS-5.}
\vspace{-15pt}
\label{fig:dialect_distribution_in_madis}
\end{figure}

\subsection{Evaluation Dataset: MADIS-5 Benchmark}
We manually curate a dataset for multi-domain ADI in speech (MADIS-5) to facilitate evaluation of cross-domain robustness of ADI systems. 
Our dataset comprises $\sim$12 hours of speech (4854 utterances) collected from four different public sources with varying similarity to the TV broadcast domain of ADI-5. 
The recordings were manually segmented and labeled by a native Arabic speaker with linguistic expertise (PhD in Computational Linguistics) and extensive exposure to Arabic language variation. 
The dialect labels were then verified by another native Arabic speaker with  competence and keen interest in different Arabic dialects. 
The data sources are:

\begin{itemize}
   \item \textbf{Radio broadcast.} Similar to previous research \cite{foley-etal-2024-geolocating}, we harvested radio broadcasts using the radio.garden website to access local radio stations throughout the Arab world during the time period December 2024 - January 2025.

   
   \item \textbf{TV dramas.} We compiled 5-7 second speech segments from the Arabic Spoken Dialects Regional Archive (SARA) on the Kaggle platform.
   
   \item \textbf{TEDx Talks.} We created short segments and added dialect annotations to the Arabic portion of the TEDx dataset \cite{salesky2021mtedx}.
   
   \item \textbf{Theater.} We used YouTube to collect Arabic dramatic and comedy plays performed in theaters in different time periods, which we then segmented into short samples.
\end{itemize}

\noindent
These sources are characterized by varying recording conditions and diverse themes, which makes the dataset ideal for evaluating cross-domain robustness of ADI systems beyond TV broadcast.
It is worth noting that for native Arabic speakers with exposure to multiple dialects through media, a coarse-grained classification of Arabic dialects is a relatively trivial task. 
This stems mainly from the abundance of discriminating linguistic features at both the phonetic and lexical levels that signal regional dialects. 
Although our two annotators achieved perfect agreement in categorizing regional dialectal speech, they disagreed on the classification of 2.3\% of radio broadcast segments as dialect or MSA. 
This disagreement reflects an ongoing research challenge, as MSA in some scenarios exists on a continuum with dialectal Arabic \cite{keleg-etal-2024-estimating, keleg-etal-2023-aldi}. 
We release the dataset on Hugging Face hub under a CC BY-NC 4.0 license. 

\section{Experimental Setup}
\label{section:setup}
To evaluate the effectiveness of voice conversion for improving ADI, we compare it against several strong text and speech baselines as well as various audio augmentation techniques.

\subsection{Text and Speech Baselines}
For text baselines, we transcribe the datasets using two publicly available models: (1) a universal phoneme recognizer for phonetic transcriptions\footnote{facebook/wav2vec2-lv-60-espeak-cv-ft}, and (2) a Whisper-based model for orthographic transcriptions\footnote{speechbrain/asr-whisper-large-v2-commonvoice-ar}. 
We then train SVM classifiers for each transcription system with hyperparameters optimized using the scikit-learn library. 
We chose SVMs because they have demonstrated superior performance over neural classifiers in discriminating closely related languages from text \cite{medvedeva2017sparse}. 
Additionally, we use the orthographic transcriptions to fine-tune a BERT model pre-trained on a mixture of MSA and Arabic dialectal \cite{inoue-etal-2021-interplay} to perform ADI\footnote{CAMeL-Lab/bert-base-arabic-camelbert-mix}. 

For speech baselines, we fine-tune multilingual pre-trained speech models based on the wav2vec2 architecture; XLSR-53 \cite{babu2021xlsr}, XLSR-128 \cite{baevski2022xlsr128}, and MMS \cite{pratap2024scaling}.
We use natural segments (10 seconds max) from the ADI-5 training dataset for fine-tuning.
While all these models have 300M parameters, they differ in their pre-training data volume, with MMS having been pre-trained on more than 1,000 languages.
We experimented with different learning rates ($1 \times 10^{-4}$, $\{1, 3, 5, 7\} \times 10^{-5}$), out of which $5 \times 10^{-5}$ was optimal for the all pre-trained models. 
We trained our models for 6 epochs using the default parameters of the Hugging Face trainer.

\subsection{Data Augmentation}
We compare our voice conversion method against several common data augmentation techniques in speech processing. 
These include SpecAugment \cite{park2019specaugment}, which applies time stretching,  time masking and frequency masking to the spectrogram; pitch shifting to alter the fundamental frequency; room impulse response (RIR) simulation to emulate different acoustic environments; and additive noise augmentation using both music and background noise \cite{musan2015} at various signal-to-noise ratios.

\subsection{Voice Conversion}

In our study, we use nearest neighbor voice conversion ($k$NN-VC), a simple yet effective method for VC that does not require text transcriptions of any sort \cite{baas23_interspeech}, making it ideal for untranscribed dialectal speech. 
k-NN VC transforms a speech segment into a target voice using only a few examples of the target speaker
as a reference. 
We experimented  with k-NN VC on Arabic speech and observed that it produces high-quality output in the target voice while preserving intelligibility and relevant cues for dialect classification. 
For our study, we used native Arabic target voices from LibriVox audio books, with approximately one minute of speech per speaker.
All models using data augmentation or VC are trained for 3 epochs on the combined natural and re-synthesized datasets.
Note that the test segments remain unmodified and were not subjected to any transform.

\section{Experiments and Results}

\begin{table}[t]
\caption{Our best model compared to the SoTA in the literature. $\Delta$ shows relative improvement over MIT-QCRI system (\%).}
\vspace{-8pt} 
\centering
\renewcommand{\arraystretch}{1.05}
\setlength{\tabcolsep}{4pt}
\begin{tabular*}{\columnwidth}{@{\extracolsep{\fill}}lcccccc@{}}
\toprule
 & \multicolumn{2}{c}{\textsc{Accuracy}} & \multicolumn{2}{c}{\textsc{Precision}} & \multicolumn{2}{c}{\textsc{Recall}} \\
\cmidrule(lr){2-3} \cmidrule(lr){4-5} \cmidrule(lr){6-7}
Model & \small \% & \small $\Delta$ & \small \% & \small $\Delta$ & \small \% & \small $\Delta$ \\
\hline
MIT-QCRI \cite{ali2017speech} & 75.0 & -- & 75.1 & -- & 75.5 & -- \\
UTD \cite{ali2017speech} & 79.8 & +6.4 & 79.9 & +6.4 & 80.3 & +6.4 \\
\hline
ResNet (R) \cite{kulkarni2023yet} & 80.4 & +7.2 & 80.4 & +7.1 & 80.5 & +6.6 \\
ECAPA (E) \cite{kulkarni2023yet} & 82.5 & +10.0 & 82.6 & +10.0 & 82.7 & +9.5 \\
Fusion (R + E) \cite{kulkarni2023yet} & 84.7 & +12.9 & 84.8 & +12.9 & 84.9 & +12.5 \\
\hline
\rowcolor{lightblue}
MMS-VC (Ours) & \textbf{85.3} & \textbf{+13.7} & \textbf{85.4} & \textbf{+13.7} & \textbf{85.3} & \textbf{+13.0} \\
\bottomrule
\end{tabular*}

\label{tab:sota_metrics_comparison}
\vspace{-15pt} 
\end{table}

\subsection{Baselines}
\textbf{Text baselines.}  On the in-domain test set, the phone-based SVM (66.82\%) outperforms both the  character-based SVM (50.00\%) and Arabic BERT (62.73\%).
This shows that text-based classifiers trained on ASR transcripts are not reliable for the ADI task since ASR models are trained on MSA speech and normalize dialect-specific lexical features in their output. 
However, the phone-based SVM performs worse than Arabic BERT on the cross-domain setting (57.90\% vs 59.40\%). 

\noindent
\textbf{Speech baselines.}  MMS with 75.94\% in-domain accuracy demonstrates clear advantages over both XLSR variants, outperforming XLSR-128 (70.58\%) and XLSR-53 (68.57\%). 
This indicates that MMS's massive multilingual pre-training provides better feature representations for dialect identification. However, all models perform poorly on out-of-domain data, with XLSR variants even falling behind two of the text baselines.  
This consistent performance drop on out-of-domain data highlights a critical limitation: fine-tuning pre-trained speech models yields ADI systems that transfer poorly to unseen domains or  ``in-the-wild" speech that differs from the training data. 
Given that MMS performance is superior compared to XLSR variants, we perform our main experiments with data augmentation and voice conversion by fine-tuning MMS. 

\subsection{ADI with Data Augmentation}

Audio augmentation techniques improve upon the MMS model trained only natural speech, with the combination of all augmentations achieving 81.64\% accuracy (+7.51\% in-domain relative improvement over MMS baseline). 
When applying a single augmentation in isolation, pitch shifting proves most effective (80.63\%, +6.18\%), followed by additive noise (79.42\%, +4.58\%). 
We also observe consistent improvements in the cross-domain setting across all data augmentation techniques. 

\subsection{ADI with Voice Conversion}
Fine-tuning MMS on a combination of natural and re-synthesized speech using VC yields the most substantial improvements in our study. 
Even with just one target speaker for VC, our approach achieves 82.17\% in-domain accuracy, surpassing all data augmentation methods. 
Increasing the number of target speakers consistently enhances performance, reaching 85.32\% with four target speakers with a 12.35\% relative improvement over the baseline MMS model. 
Additionally, we observe consistent cross-domain improvements with the model trained on four target voices achieves state-of-the-art results across all domains (overall relative improvement of up to 34.07\%). 
Our best model shows exceptional performance on radio (87.86\%) and TEDx  (86.73\%) domains that matches in-domain performance. 
The strong cross-domain performance suggests that voice conversion is indeed an effective strategy for learning representations that are robust to domain shifts.

\subsection{Comparison to state-of-the-art on ADI-5}

The ADI-5 dataset serves as a standard benchmark for Arabic dialect identification in the literature \cite{ali2017speech, kulkarni2023yet}. 
Here, we compare the in-domain performance of our model to previously reported results as illustrated in Table \ref{tab:sota_metrics_comparison}. 
Our model, trained with re-synthesized speech using voice conversion, sets a new state-of-the-art performance, outperforming all previous approaches across all metrics. 
Even the strong ECAPA-TDNN system trained on SSL representations and its fusion with ResNet architecture falls short of our approach by at least 0.5\% across all metrics. 
While the fusion of ResNet and ECAPA models demonstrates the benefits of model combination (84.7\%), our single model still outperforms this ensemble approach (85.3\%). 
These results provide convincing evidence of the effectiveness of voice conversion for ADI.

\begin{table}[t]
\caption{Model accuracy in unbiased and biased conditions. The $\Delta$ columns show relative improvement over MMS baseline.}
\vspace{-8pt} 
\centering
\setlength{\tabcolsep}{4pt}
\begin{tabular*}{\columnwidth}{@{\extracolsep{\fill}}lcccc@{}}
\toprule
& \multicolumn{2}{c}{In-domain} & \multicolumn{2}{c}{Cross-domain} \\
\cmidrule(lr){2-3} \cmidrule(lr){4-5}
Model & \small ADI-5 & \small $\Delta$ & \small Avg. & \small $\Delta$ \\
\midrule
MMS & 75.94 & -- & 59.60 & -- \\
MMS-VC (unbiased) & \textbf{83.38} & \textbf{$+$09.80} & \textbf{76.61} & \textbf{$+$28.54} \\
MMS-VC (biased)  & 27.33 &  $-$64.01 & 24.32 & $-$59.19 \\
\bottomrule
\end{tabular*}
\label{tab:results_comparison}
\vspace{-15pt} 
\end{table}

\section{Model Analysis}

In the previous section, we established that voice conversion is an  effective method for training robust ADI systems.
Here, we investigate why voice conversion yields such substantial improvements and better cross-domain generalizations. 
Our hypothesis is that re-synthesizing the training data using voice conversion helps normalize speaker variations in the dataset, which otherwise introduce a significant bias. 
This bias stems from an inherent limitation in dialect identification datasets: each speaker typically speaks only one dialect at the native level, therefore the training segments for each dialect are drawn from a disjoint speaker set.
This lack of speaker overlap between dialects creates a strong association between speaker identity and dialect label. 
Consequently, neural networks can exploit speaker identity as an easier shortcut for dialect classification, rather than learning the more subtle but relevant dialectal features.
While collecting training samples from a large native speaker pool could mitigate this bias, there is still no explicit incentive that prevents the model from simply memorizing the speakers rather than learning robust dialect representations.

To test this hypothesis, we conduct a controlled experiment by training models on re-synthesized speech in two conditions: unbiased and biased.
In the unbiased setting, we re-synthesize the training speech samples using a unified set of 12 target speakers across all dialects, ensuring no association between the speaker and the spoken dialect (i.e., target speakers for VC are uniformly distributed across dialects).
This setting is similar to our best performing model, but excludes the natural speech data from the final training dataset.
In the biased setting, we deliberately introduce speaker bias by using dialect-specific target speakers: we re-synthesize the training data using a pool of 60 target speakers, with a disjoint set of 12 speakers for each dialect.
This setting resembles the natural training data but with limited speaker variation within each dialect.
As in our previous experiment, the evaluation samples for this experiment are not modified and remain in their natural form. 
The results of this experiment, shown in Table~\ref{tab:results_comparison}, are quite surprising.
Fine-tuning MMS on re-synthesized speech only (no natural speech) in the unbiased condition yields substantial improvements, achieving 83.38\% in-domain accuracy (+9.80\%) and, more importantly, 76.61\% cross-domain accuracy (+28.54\%).
In contrast, the accuracy of the model trained on the biased re-synthesized dataset drops dramatically to 27.33\% in-domain and 24.32\% cross-domain, performing close to random chance.
These results strongly support our hypothesis that neural networks indeed exploit speaker-predictive features when available, and that voice conversion can effectively reduce this bias, leading to more robust dialect identification systems.

\section{Discussion and Conclusion}

Our experiments demonstrated that voice conversion significantly improves the generalization of ADI systems, particularly in cross-domain scenarios. 
The remarkable performance gain achieved through voice conversion (+34.07\% cross-domain) cannot be matched by traditional data augmentation techniques. 
Our analysis with controlled experiments revealed that voice conversion is effective because it eliminates the speaker bias in ADI datasets, which current models are likely to exploit as shortcuts for dialect classification.

When we initiated this research, we hypothesized that voice conversion could serve as a data augmentation technique to expand our training data.
However, our investigation revealed that the benefits of voice conversion extend far beyond simple data augmentation as it serves as an effective technique for bias mitigation in language and dialect identification datasets.
This insight opens up new avenues for handling dataset biases in speech technology as we demonstrate the value of taking a data-centric approach to model development, where  understanding  and mitigating dataset biases can lead to more substantial improvements than model refinements alone.
Furthermore, our proposed approach could be extended to other tasks where speaker identity is strongly correlated with output labels, such as accent identification and speech classification for healthcare applications.

\section{Acknowledgements}
We thank the anonymous reviewers for their positive feedback. 
We sincerely thank Aravind Krishnan for his valuable comments on the work presented in this paper.
This research is funded by the Deutsche Forschungsgemeinschaft (DFG, German Research Foundation), Project-ID 232722074 -- SFB 1102.

\bibliographystyle{IEEEtran}
\bibliography{paperbib}

\end{document}